\documentclass[letterpaper, 10 pt, conference]{ieeeconf}
\IEEEoverridecommandlockouts
\usepackage{amsmath,amsfonts}
\usepackage{algorithmic}
\usepackage{algorithm}
\usepackage{array}
\usepackage[caption=false,font=normalsize,labelfont=sf,textfont=sf]{subfig}
\usepackage{textcomp}
\usepackage{stfloats}
\usepackage{url}
\usepackage{verbatim}
\usepackage{graphicx}
\usepackage{cite}
\usepackage{amsfonts}
\usepackage{amsmath}
\usepackage{booktabs}
\usepackage{multirow}
\usepackage{tabularx}
\let\labelindent\relax
\usepackage{enumitem}
\usepackage{xcolor}
\definecolor{skyblue}{RGB}{80,160,205}
\usepackage[colorlinks=true, urlcolor=skyblue]{hyperref}

\AtBeginDocument{
  \setlength{\abovedisplayskip}{4pt plus 1pt minus 1pt}
  \setlength{\belowdisplayskip}{4pt plus 1pt minus 1pt}
  \setlength{\abovedisplayshortskip}{2pt plus 1pt minus 1pt}
  \setlength{\belowdisplayshortskip}{3pt plus 1pt minus 1pt}
  \setlength{\jot}{2pt}
}

\begin{document}

\title{\LARGE \bf Online Sim-to-Real Adaptation via Closed-Loop System Modeling}

\author{Yuhao Huang$^{1}$, Samuel A. Moore$^{1}$, and Boyuan Chen$^{1}$%
\thanks{This work is supported by DARPA FoundSci program under award HR00112490372, DARPA TIAMAT program under award HR00112490419, ARO under award W911NF2410405, ARL STRONG program under awards W911NF2320182, W911NF2220113, and W911NF242021. $^1$ All authors are from Duke University.}
}

\maketitle
\bstctlcite{IEEEexample:BSTcontrol}

\begin{abstract}
Sim-to-real transfer has made substantial progress, but can still produce controllers that remain stable and functional on hardware while suffering from degraded tracking accuracy due to residual dynamics mismatch. Correcting these errors typically requires identifying the underlying system dynamics, adapting the control policy, or returning to simulation for additional training and finetuning, all of which can require substantial data and computation. We propose OSRAM (Online Sim-to-Real Adaptation via Closed-Loop System Modeling), a framework that instead adapts the reference commands provided to an existing controller. OSRAM treats the deployed robot and its policy as a unified closed-loop dynamical system and learns its task-level command-response behavior directly from tracking observations. A closed-loop dynamics model is meta-trained across randomized dynamics in simulation and rapidly finetuned after deployment using limited real-world interaction. The adapted model is then used to optimize future reference commands while leaving the underlying control policy unchanged. We evaluate OSRAM on bipedal velocity tracking and loco-manipulation in simulation and on hardware. Results show that closed-loop modeling improves prediction and tracking accuracy under unseen dynamics, while online reference adaptation reduces residual sim-to-real tracking errors across different control objectives and hardware configurations. These results demonstrate that adapting the behavior of the robot-policy closed loop provides a practical alternative to finetuning the policy or identifying the full physical dynamics for sim-to-real transfer. More information can be found at \url{http://generalroboticslab.com/OSRAM}.



\end{abstract}


\section{Introduction}
Sim-to-real transfer has enabled learning-based controllers to execute increasingly dynamic and complex behaviors on physical robots~\cite{sim2real_xie, sim2real_manipulator}. Yet even when a policy transfers successfully and remains functional on hardware, its execution can suffer from degraded precision due to residual mismatch between simulated and real-world dynamics\cite{sim2real_survey}. These discrepancies can arise from modeling inaccuracies, complex actuator dynamics, sensor noise, latency, and environmental variations that are difficult to capture fully in simulation. Even modest mismatches can produce persistent tracking errors in dynamic and contact-rich tasks~\cite{motor_adaptation}.

Existing approaches often address the sim-to-real gap by improving policy robustness during training. Domain randomization~\cite{domain_randomization,dynamics_randomization} varies physical parameters such as mass, friction, latency, and actuator dynamics in simulation, while related methods inject sensor, action, or adversarial perturbations. These methods rely on the real system being sufficiently represented by the training distribution. Designing this distribution can be difficult in practice. Insufficient randomization leaves the policy vulnerable to unseen dynamics, whereas overly broad randomization can sacrifice task performance or destabilize learning~\cite{domain_randomization_drawback}. Consequently, residual tracking errors can remain after deployment even when the policy remains functional.

\begin{figure}
    \centering
    \includegraphics[width=\columnwidth]{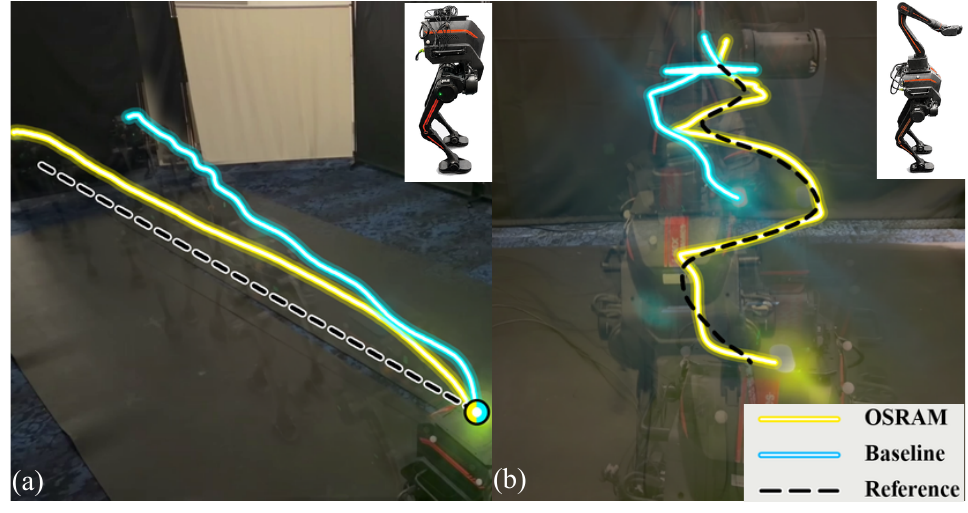}
    \vspace{-20pt}
    \caption{OSRAM reduces residual sim-to-real tracking errors through online reference-command adaptation. (a) Velocity tracking without position feedback. (b) Loco-manipulation sine wave trajectory tracking.}
    \label{fig:motivation}
    \vspace{-20pt}
\end{figure}

Another line of work adapts after deployment by estimating physical parameters or learning representations of real-world dynamics from interaction data~\cite{spi_active,asap}. These estimates can be used to update the simulator and retrain or fine-tune the policy. Although effective, such approaches seek to recover aspects of the underlying physical system and can require substantial real-world data, simulator calibration, and additional optimization. For high-dimensional robotic systems, accurately identifying the dynamics can itself be challenging, while modifying a deployed policy introduces additional computational and validation costs.

In this work, we focus on a complementary regime where the transferred policy remains functional but does not track its intended commands with sufficient accuracy. Instead of reconstructing the physical dynamics or modifying the policy, our key idea is to correct residual sim-to-real errors by adapting the \textit{reference commands} sent to the existing controller (Fig.~\ref{fig:motivation}). An appropriately modified reference can compensate for systematic under-tracking, overshoot, or delay while leaving the underlying controller unchanged.

To determine how these references should be modified, we treat the robot and its policy together as a closed-loop dynamical system and learn its task-level command-response behavior directly from tracking observations. This formulation shifts adaptation from reconstructing high-dimensional physical dynamics to modeling the lower-dimensional behavior directly relevant to the task. Because the learned model captures the aggregate response of both the robot and its embedded controller, it can predict how candidate reference commands will actually be executed.

Building on this insight, we introduce OSRAM, a framework for online reference-command adaptation through closed-loop system modeling. OSRAM meta-trains a closed-loop dynamics model across randomized system configurations in simulation, rapidly fine-tunes it from limited interaction after deployment, and uses the adapted model to optimize future reference commands. OSRAM leaves the deployed policy unchanged and does not require explicit identification of physical parameters. We evaluate OSRAM on bipedal velocity tracking and loco-manipulation in simulation and on hardware, demonstrating improved prediction and tracking under unseen dynamics and across different control objectives and hardware configurations.

\section{Related Work}

\subsection{Sim-to-Real Transfer with Online Adaptation}

Online adaptation complements robustness-oriented sim-to-real methods by using interaction data collected after deployment to adjust system behavior. A common strategy is to estimate latent environment representations or physical parameters from online observations and condition the policy on these estimates~\cite{yu2017preparing,kumar2021rma,kumar2022adapting,prime2026}. For example, UPOSI~\cite{yu2017preparing} combines a universal policy with online system identification, whereas Rapid Motor Adaptation (RMA)~\cite{kumar2021rma} infers latent environment representations from observation histories to rapidly adapt locomotion across varying terrains and dynamics. These approaches achieve effective online adaptation, but generally require additional latent inference, observation encoding, or system identification modules. In contrast, OSRAM leaves the deployed policy unchanged and adapts its reference commands through a learned model of the closed-loop command-response behavior.

Another line of work focuses on residual or behavior-level adaptation, where corrective residual policies or behavior adaptation modules are learned from real-world interaction data or updated online~\cite{handelbot,zhang2023efficient,chi_iterative,loco_residual}. Online behavior-centric adaptation~\cite{loco_residual} adapts bipedal locomotion by aligning behavior representations and adjusting low-level controller parameters under unmodeled dynamics mismatch. Sym2Real~\cite{sym2real} learns symbolic dynamics from simulation and uses limited real-world data to learn residual corrections for data-efficient adaptive control. These methods demonstrate that real-world adaptation can be achieved without retraining a controller from scratch, but perform adaptation by modifying the controller, its actions, or its underlying dynamics representation. OSRAM instead performs adaptation entirely through the reference interface. It models the deployed robot-policy closed loop and optimizes future reference commands without modifying the policy or its parameters.

\subsection{Input-Output Dynamics Modeling for Robotics}

Dynamics models are widely used for system identification, policy learning, and model-based control~\cite{chua2018deep,pilco,sns}. A common formulation models the underlying system through state transitions,
\begin{equation}
\mathbf{x}_{t+1} = f(\mathbf{x}_t, \mathbf{u}_t),\quad
\mathbf{y}_t = h(\mathbf{x}_t, \mathbf{u}_t),
\end{equation}
where $\mathbf{x}_t \in \mathcal{X} \subseteq \mathbb{R}^n$ is the system state, $\mathbf{u}_t \in \mathcal{U} \subseteq \mathbb{R}^m$ is the control input, and $\mathbf{y}_t \in \mathcal{Y} \subseteq \mathbb{R}^p$ is the system output. Learned state-space models can represent complex nonlinear dynamics, with recent approaches using recurrent~\cite{dreamer3} and transformer-based~\cite{levy2026simulation} architectures to capture longer-horizon behavior. However, learning sufficiently accurate state transitions for high-dimensional robotic systems can require substantial data, particularly when only a low-dimensional aspect of the resulting behavior is relevant to the downstream task success.

Input-output system identification provides an alternative that predicts observable responses directly from histories of inputs and outputs. For example, NARX~\cite{joumah2024forward,telli2023quadrotor} and behavioral system models~\cite{ref_steer,willems2007behavioral} describe robotic systems implicitly:
\begin{equation}
\mathbf{y}_{t+1} = F(\mathbf{u}_{t-N:t}, \mathbf{y}_{t-N:t}),
\end{equation}
avoiding explicit reconstruction of the complete internal state. This perspective is particularly useful for partially observed systems or when only task-relevant outputs need to be predicted. Most closely related to our work, Zeng et al.~\cite{ref_steer} introduce data-driven reference steering, which treats the reference trajectory as the input and the realized trajectory as the output of an existing closed-loop system, and uses data-driven predictive control to modify reference trajectories online while preserving the underlying controller.

OSRAM shares this reference-level adaptation but addresses a different setting. Instead of relying on an approximately linear closed-loop representation constructed from offline input-output data, OSRAM learns a nonlinear closed-loop command-response model. The model is meta-trained across randomized dynamics in simulation and rapidly adapted from limited deployment data, enabling adaptation to unseen sim-to-real discrepancies. Our method enables reference-level adaptation for learning-based controllers and nonlinear closed-loop dynamics.



\section{Method}
\label{sec:closed-loop model}
\begin{figure*}[t]
    \centering
    \includegraphics[width=1\textwidth]{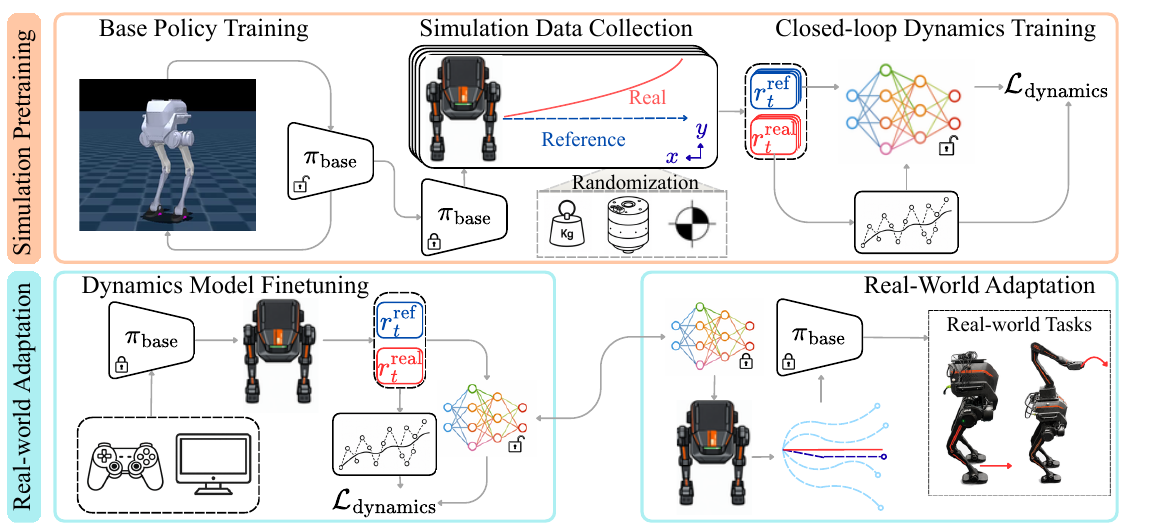}
    \vspace{-20pt}
    \caption{Overview of OSRAM. During simulation pretraining, a base policy is trained and used to collect closed-loop rollouts under randomized system dynamics, from which a meta-learned command-response model is trained. After deployment, the model is rapidly finetuned using limited real interaction and used to optimize future reference commands. The underlying control policy remains unchanged throughout adaptation.}
    \label{fig:pipeline}
    \vspace{-20pt}
\end{figure*}

OSRAM adapts a deployed robotic system through its reference interface without modifying the underlying control policy. The framework consists of two stages (Fig.~\ref{fig:pipeline}). First, we learn a closed-loop command-response model in simulation and meta-train it across randomized system dynamics to enable rapid adaptation. Second, after deployment, we finetune this model using real-world tracking observations and use it to optimize future reference commands.

\noindent\textbf{Closed-Loop Command-Response Modeling.} Consider a deployed policy $\pi$ that receives a task-level reference command $\mathbf{r}^{\mathrm{ref}}_t$ and produces low-level actions for the robot. Rather than separately modeling the physical robot dynamics and the policy, we treat their composition as a single closed-loop dynamical system. The resulting behavior can then be characterized directly through the relationship between commanded references and realized responses.


Specifically, we learn a model $F_\theta$ that predicts the next realized task-level response from a history of reference commands and observed responses:
\begin{equation}
\hat{\mathbf{r}}^{\text{real}}_{t+1} 
= F_\theta\!\left(
\mathbf{r}^{\text{real}}_{t-N:t},\ 
\mathbf{r}^{\text{ref}}_{t-N:t}
\right),
\end{equation}
where $\mathbf{r}^{\text{ref}}_t \in \mathbb{R}^b$ denotes the reference command sent to the policy at time $t$, $\mathbf{r}^{\text{real}}_t \in \mathbb{R}^b$ denotes the observed tracking response, and $\hat{\mathbf{r}}^{\mathrm{real}}_{t+1}$ denotes the predicted response.

This formulation captures the aggregate behavior of the robot and its embedded controller without requiring access to the policy internals or an explicit model of the robot's physical dynamics. It also restricts modeling to task-relevant variables, which are often substantially lower dimensional than the full robot state. As a result, adaptation can focus directly on the command-response discrepancies that affect task execution.

\noindent\textbf{Closed-Loop Meta-Dynamics Training.} To enable rapid adaptation to previously unseen dynamics, we meta-train the closed-loop command-response model $F_\theta$ using a Reptile-inspired first-order meta-learning procedure~\cite{reptile}. The goal is to learn a model prior that can rapidly adapt to previously unseen system dynamics using only a small amount of target-domain data. Meta-training is performed entirely in a high-fidelity simulation environment using self-supervised trajectory data, without access to real-world data during pretraining.

During data collection, key physical parameters of the simulator, including mass, friction, and actuator dynamics, are randomized to cover a broad range of in-distribution variations. Stochastic excitation signals are also injected into the reference commands to excite diverse operating regimes and dynamic responses.

Each randomized system configuration defines a meta-training task $\mathcal{D}_m$ containing trajectories of reference commands and realized responses. Starting from a shared initialization, the model is adapted for several inner-loop gradient steps on each sampled task. The shared parameters are then moved toward the corresponding task-specific solutions. Repeating this process across diverse simulated dynamics produces an initialization that can be rapidly specialized using limited target-domain data.

We train the model using a combination of one-step prediction loss and multi-step rollout consistency:
\begin{equation}
\begin{aligned}
\mathcal{L}_{\mathrm{dynamics}}
&=
\underbrace{\lambda_0 
\left\lVert 
\mathbf{r}_{t+1}^{\mathrm{real}} 
- F_\theta\!\left(
\mathbf{r}^{\mathrm{real}}_{t-N:t},\ 
\mathbf{r}^{\mathrm{ref}}_{t-N:t}
\right)
\right\rVert_2^2}_{\mathcal{L}_{\mathrm{step}}}
\\[-0.6em] \quad+
&\underbrace{\lambda_1 \frac{1}{K} \sum_{k=1}^{K} \gamma^k 
\left\lVert 
\mathbf{r}_{t+k}^{\mathrm{real}} 
- F_\theta^{(k)}\!\left(
\mathbf{r}^{\mathrm{real}}_{t-N:t},\ 
\mathbf{r}^{\mathrm{ref}}_{t-N:t+k-1}
\right)
\right\rVert_2^2}_{\mathcal{L}_{\mathrm{rollout}}}.
\end{aligned}
\end{equation}
where $F_\theta^{(k)}(\cdot)$ is a $k$-step autoregressive rollout of the model, obtained by recursively feeding predicted response back into the model. The discount factor $\gamma \in (0,1]$ reduces the contribution of errors farther into the rollout, while $\lambda_0$ and $\lambda_1$ balance one-step accuracy and long-horizon consistency.

The resulting meta-learned model provides a prior over closed-loop command-response behavior. After deployment, it can be finetuned directly from observed reference-response trajectories without modifying the policy or reconstructing the underlying physical dynamics. The overall procedure is summarized in Algorithm~\ref{alg:meta_dynamics}.

\begin{algorithm}[t]
\caption{Closed-Loop Meta-Dynamics Training \& Adaptation}
\label{alg:meta_dynamics}
\begin{algorithmic}[1]
    \STATE \textbf{Require:} Simulation task distribution $\mathcal{D}_{\mathrm{sim}}$, update steps $N_{\mathrm{in}}$, inner learning rate $\alpha$, meta learning rate $\beta$, finetuning steps $J$, number of tasks per update $M$
    \STATE Initialize shared  model parameters $\theta_0$
    \STATE \texttt{// Simulation meta-training}
        \WHILE{not converged}
        \STATE Sample tasks $\{\mathcal{D}_m\}_{m=1}^{M} \sim \mathcal{D}_{\mathrm{sim}}$
        \FOR{$m=1,\ldots,M$}
            \STATE $\theta^{(m)}_0 \leftarrow \theta$
            \FOR{$n=1,\ldots,N_{\mathrm{in}}$}
                \STATE $\theta^{(m)}_n
                \leftarrow
                \theta^{(m)}_{n-1}
                -
                \alpha
                \nabla_{\theta}
                \mathcal{L}_{\mathrm{dynamics}}
                (\mathcal{D}_m,\theta^{(m)}_{n-1})$
            \ENDFOR
        \ENDFOR
        \STATE $\theta
        \leftarrow
        \theta
        +
        \beta
        \frac{1}{M}
        \sum_{m=1}^{M}
        \left(
        \theta^{(m)}_{N_{\mathrm{in}}}
        -
        \theta
        \right)$
    \ENDWHILE
    \STATE \texttt{// Real-world specialization}
    \STATE Collect reference-response data $\mathcal{D}_{\mathrm{real}}$
    \FOR{$j=1,\ldots,J$}
        \STATE $\theta
        \leftarrow
        \theta
        -
        \alpha
        \nabla_{\theta}
        \mathcal{L}_{\mathrm{dynamics}}
        (\mathcal{D}_{\mathrm{real}},\theta)$
    \ENDFOR
\end{algorithmic}
\end{algorithm}

\noindent\textbf{Online Reference-Command Adaptation.} After deployment, OSRAM collects a small amount of reference-response data and finetunes the meta-learned closed-loop model to the deployed system. The adapted model is then used to optimize future reference commands so that the resulting robot behavior more closely follows the original desired trajectory.

Let $\mathbf{r}^{\mathrm{ref}}_{0:T}$ denote the nominal reference trajectory and $\mathbf{r}^{\mathrm{ref,new}}_{0:T}$ a candidate adapted trajectory. We optimize the latter using the learned closed-loop model:
\begin{equation}
\begin{aligned}
\mathbf{r}^{\mathrm{ref},*}_{0:T}
=
\arg\min_{\mathbf{r}^{\mathrm{ref,new}}_{0:T}}
\sum_{t=0}^{T}\gamma^t
\Big[
&\ell_{\mathrm{track}}
\big(
\hat{\mathbf{r}}^{\mathrm{real}}_t,
\mathbf{r}^{\mathrm{ref}}_t
\big)
\\
+&\ell_{\mathrm{dev}}
\big(
\mathbf{r}^{\mathrm{ref,new}}_t,
\mathbf{r}^{\mathrm{ref}}_t
\big)
\\
+&\ell_{\mathrm{smooth}}
\big(
\mathbf{r}^{\mathrm{ref,new}}_t,
\mathbf{r}^{\mathrm{ref,new}}_{t-1}
\big)
\Big]
\\
\mathrm{s.t.}\quad
&\hat{\mathbf{r}}^{\mathrm{real}}_{t+1}
=
F_\theta\!\left(
\hat{\mathbf{r}}^{\mathrm{real}}_{t-N:t},
\mathbf{r}^{\mathrm{ref,new}}_{t-N:t}
\right).
\end{aligned}
\label{eq:reference_optimization}
\end{equation}

Here, $\gamma \in [0,1]$ is a discount factor that mitigates the accumulation of model errors over long planning horizons. For the initial prediction window, observed responses are used wherever available, and missing history is filled using a buffer of previously executed references and responses. The tracking term $\ell_{\mathrm{track}}$ encourages the predicted realized response to follow the original desired reference. The deviation term $\ell_{\mathrm{dev}}$ regularizes the adapted command toward the nominal reference, preventing unnecessarily aggressive corrections and reducing extrapolation beyond the model's observed operating regime. The smoothness term $\ell_{\mathrm{smooth}}$ penalizes abrupt temporal changes in the adapted reference. These terms allow OSRAM to compensate for systematic tracking bias, delay, and other residual sim-to-real discrepancies while preserving the behavior induced by the original controller.

Importantly, $\mathbf{r}^{\mathrm{ref,new}}$ is not a replacement task objective. It is an intermediate command selected so that the realized response $\mathbf{r}^{\mathrm{real}}$ more accurately follows the original desired reference $\mathbf{r}^{\mathrm{ref}}$. Thus, adaptation occurs entirely through the reference interface, while the underlying control policy remains fixed.

\noindent\textbf{Sampling-Based Reference Optimization.} We solve Eq.~\eqref{eq:reference_optimization} using model predictive path integral (MPPI) control, which efficiently evaluates large numbers of candidate reference trajectories through parallel rollouts of the learned model. At each planning step, candidate reference sequences are sampled around the current nominal trajectory, evaluated according to Eq.~\eqref{eq:reference_optimization}, and combined according to their trajectory costs.

Given $K$ sampled candidate reference trajectories $\{\mathbf{r}^{\mathrm{ref,sample}}_{0:T}(k)\}_{k=1}^{K}$ with corresponding costs $\mathcal{R}(k)$, we compute
\begin{equation}
w_k
=
\frac{
\exp\!\left(
-\frac{
\mathcal{R}(k)-\mathcal{R}_{\min}
}{
\lambda
\left(
\mathcal{R}_{\max}-\mathcal{R}_{\min}
\right)
}
\right)
}{
\sum_{j=1}^{K}
\exp\!\left(
-\frac{
\mathcal{R}(j)-\mathcal{R}_{\min}
}{
\lambda
\left(
\mathcal{R}_{\max}-\mathcal{R}_{\min}
\right)
}
\right)
},
\end{equation}
and update the reference trajectory as
\begin{equation}
\mathbf{r}^{\mathrm{ref,new}}_{0:T}
=
\sum_{k=1}^{K}
w_k\,
\mathbf{r}^{\mathrm{ref,sample}}_{0:T}(k).
\label{eq:mppi_update}
\end{equation}
The temperature parameter $\lambda$ controls the concentration of the weighting over candidate trajectories. We iteratively repeat the sampling and weighting procedure while annealing the sampling covariance following past work~\cite{dial-mpc}. The first portion of the optimized reference is executed, new observations are incorporated, and the optimization is repeated in a receding-horizon manner.

\section{Experimental Results}
\label{sec:result}
We evaluated OSRAM on bipedal velocity tracking and loco-manipulation in simulation and on hardware. Our experiments were designed to examine the following questions:
\begin{itemize}[label={}, leftmargin=0pt,itemindent=0pt]
    \item $\textbf{Q1}$: Does meta-dynamics learning improve adaptation performance under previously unseen out-of-distribution environment dynamics?
    \item $\textbf{Q2}$: Does modeling the closed-loop system dynamics lead to lower prediction error and higher-quality adaptation?
    \item $\textbf{Q3}$: Does the proposed framework outperform existing online adaptation-based sim-to-real transfer methods?
    \item $\textbf{Q4}$: Does the proposed framework work for sim-to-real transfer problems with different control objectives, policy configurations, and command-response spaces?
\end{itemize}

\subsection{Experimental Setup}
We briefly describe the robotic platform, task configurations, and simulation and hardware setups used to evaluate OSRAM.

\noindent\textbf{Platforms and Tasks.} We evaluated OSRAM on the LimX Tron1 robot with and without its loco-manipulation expansion kit. To assess generality across different control objectives and embodiments, we considered two tasks: base velocity tracking and loco-manipulation. Detailed task definitions and reward specifications are provided in Table~\ref{tab:task_reward}. For the loco-manipulation task, an end-effector pose-tracking objective was added to the velocity-tracking reward. All reinforcement learning policies were trained in mjlab~\cite{zakka2026mjlab} using Proximal Policy Optimization (PPO) to optimize cumulative rewards.

\begin{table}[h]
\centering
\small
\setlength{\tabcolsep}{8pt}
\renewcommand{\arraystretch}{0.95}
\caption{Task Rewards}
\vspace{-10pt}
\label{tab:task_reward}
\begin{tabular}{@{}lrlr@{}}
\toprule
\textbf{Term} & \textbf{Weight} & \textbf{Term} & \textbf{Weight} \\
\midrule
\multicolumn{4}{c}{\textbf{End-Effector(EE) Position Tracking}} \\
\midrule
EE Position & 2.0 & EE Orientation & 1.5 \\
\midrule
\multicolumn{4}{c}{\textbf{Base Velocity Tracking}} \\
\midrule
Linear velocity & 2.0 & Angular velocity & 2.0 \\
\midrule
\multicolumn{4}{c}{\textbf{Behavior Regularization}} \\
\midrule
Projected gravity         & 1.0   & Pose deviation      & 1.0 \\
DoF position limit        & -1.0  & Action rate         & -1.0 \\
Soft landing              & -1e-5 & Self collision      & -1.0 \\
Foot slip                 & -0.1  & Foot swing height   & -0.25 \\
Foot clearance            & -2.0  & Foot distance       & -2.0 \\
\bottomrule
\end{tabular}
\vspace{-17pt}
\end{table}

\noindent\textbf{Simulation and Hardware Setup.} To support efficient comparisons and ablation studies, we developed a real-time asynchronous MuJoCo~\cite{todorov2012mujoco} simulation environment in which all planners, state estimators, and control policies were implemented within the ROS~control~\cite{ros_control} framework. Because base linear velocities were required for both velocity tracking and loco-manipulation, an Invariant Extended Kalman Filter (InEKF)~\cite{hartley2020IEKF} was used in simulation to estimate base linear velocity and orientation, while joint states and global base position were read from the simulator.

For the hardware experiments, we reused the same setup to maintain consistency between simulation and real-world evaluation, while using a motion capture system for global base position estimation. To further increase the sim-to-real gap, we added extra mass and modified the control policy gains to create dynamics mismatch.

The model trainer and MPPI planner ran asynchronously in separate threads on a single RTX4090 GPU. The MPPI planner operates at 50\,Hz, with each iteration initialized using the provided reference trajectory as the nominal guess. The dynamics model trainer updates the model at 2\,Hz with a batch size of $64$. We used 30 finetuning gradient steps for velocity tracking and 70 for loco-manipulation.

\noindent\textbf{Evaluation Metrics.} We evaluated tracking performance and reliability using four metrics. The mean tracking error $E_{\text{mean}}$ captured systematic over- or under-tracking of the command. The RMSE $E_{\text{RMSE}}$ quantified the overall tracking-error magnitude, penalizing large deviations while reflecting both bias and transient fluctuations. The RMSE standard deviation $E_{\text{std}}$ was computed over repeated trials ($20$ in simulation and $5$ in hardware) and measured run-to-run consistency, with lower values indicating more repeatable behavior. Finally, the success rate $SR$ was the fraction of trials completed without falling or violating environment-terminating safety limits. All tracking-error metrics were averaged over successful trials only, whereas $SR$ was evaluated over all trials.

\subsection{Adaptation to Unseen Dynamics}
\begin{figure}[t]
    \centering
    \includegraphics[width=\columnwidth]{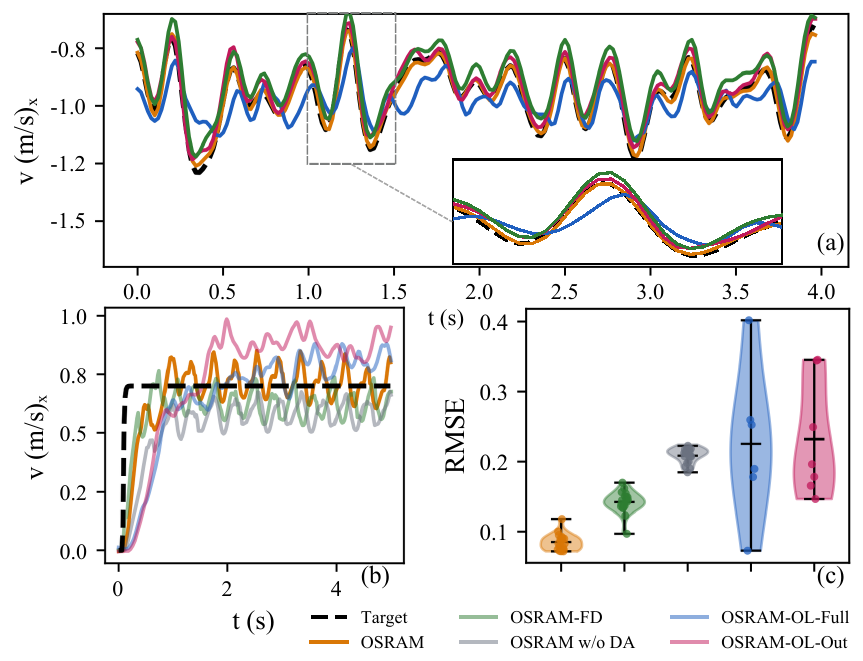}
    \vspace{-20pt}
    \caption{Model prediction and velocity-tracking comparison. (a) Predicted velocity trajectories from different dynamics models, with a magnified view highlighting local prediction differences. (b) Closed-loop velocity-tracking performance compared with the target command. (c) Distribution of tracking RMSE over multiple trials. Lower RMSE and tighter distributions indicate more accurate and consistent tracking.}
    \label{fig:method_comparison}
    \vspace{-20pt}
\end{figure}
To address \textbf{Q1}, we evaluated the prediction accuracy and planning performance of the proposed meta-dynamics model against a standard \textbf{forward MLP model (FD)}. Both models were pre-trained on the same amount of task data and then finetuned on an out-of-distribution dataset using batch size $64$ and $10$ gradient steps for a fair comparison.

We conducted two complementary evaluations. First, we compared the prediction capabilities of the two models by measuring prediction errors on out-of-distribution trajectories, where the center of mass and actuator gains of the robot were further randomized. As shown in Fig.~\ref{fig:method_comparison}(a), our meta-dynamics model achieved substantially lower prediction error in this out-of-distribution setting. For one-step prediction, our proposed model obtained a mean absolute error (MAE) of $0.0116$, compared with $0.0594$ for the forward dynamics model, corresponding to an $80.5\%$ reduction in prediction error (approximately $5\times$ lower MAE).

Second, we evaluated task-level performance by deploying both models in velocity-tracking tasks with different commanded forward velocities. This evaluation tested whether improved prediction accuracy translates into better control performance. Fig.~\ref{fig:method_comparison}(b) shows that both the forward MLP model and the meta-dynamics model improved closed-loop performance when integrated with the proposed method. The meta-dynamics model achieved the strongest tracking performance, indicating that higher prediction accuracy contributed to better task-level adaptation. This trend was further quantified in Tab.~\ref{tab:velocity_tracking_comparison}. Compared with OSRAM w/o DA, integrating a forward MLP dynamics model reduced $E_{\mathrm{RMSE}}$ from $0.1593$, $0.2127$, and $0.2209$ to $0.0800$, $0.1472$, and $0.1917$ at target velocities of $0.4$, $0.7$, and $0.9\,\mathrm{m/s}$, respectively, while maintaining a $100\%$ success rate. Replacing the forward MLP with our closed-loop meta-dynamics model further improved performance, achieving the lowest $E_{\mathrm{RMSE}}$ values of $0.0733$, $0.0929$, and $0.1560$ across the three velocity settings. OSRAM also obtained the best $E_{\mathrm{mean}}$ at all tested velocities, indicating more accurate average tracking.

\begin{table*}[t]
\centering
\caption{Ablation and Baseline Studies in Simulation - Tracking Error}
\vspace{-5pt}
\label{tab:velocity_tracking_comparison}

{
\scriptsize
\setlength{\tabcolsep}{2.4pt}
\renewcommand{\arraystretch}{0.96}
\resizebox{\textwidth}{!}{%
\begin{tabular}{@{}lcccccccccccc@{}}
\toprule
& \multicolumn{4}{c}{$v_x = 0.4\,\mathrm{m/s}$}
& \multicolumn{4}{c}{$v_x = 0.7\,\mathrm{m/s}$}
& \multicolumn{4}{c}{$v_x = 0.9\,\mathrm{m/s}$} \\
\cmidrule(lr){2-5}
\cmidrule(lr){6-9}
\cmidrule(l){10-13}
\textbf{Method}
& $E_{\mathrm{mean}}$ & $E_{\mathrm{RMSE}}$ & $E_{\mathrm{std}}$ & $SR(\%)$
& $E_{\mathrm{mean}}$ & $E_{\mathrm{RMSE}}$ & $E_{\mathrm{std}}$ & $SR(\%)$
& $E_{\mathrm{mean}}$ & $E_{\mathrm{RMSE}}$ & $E_{\mathrm{std}}$ & $SR(\%)$ \\
\midrule

\multicolumn{13}{@{}l}{\textbf{Ablation}} \\
OSRAM w/o DA
& -0.1370 & 0.1593 & \textbf{0.0064} & 100
& -0.1604 & 0.2127 & \textbf{0.0099} & 100
& -0.1292 & 0.2209 & 0.0250 & 100 \\

OSRAM-FD
& -0.0447 & 0.0800 & 0.0103 & 100
& -0.0977 & 0.1472 & 0.0152 & 100
& -0.1139 & 0.1917 & 0.0042 & 100 \\

OSRAM-OL-Full
& -0.2759 & 0.3266 & 0.0638 & 20
& -0.1017 & 0.2878 & 0.1100 & 30
& -- & -- & -- & 0 \\

OSRAM-OL-Out
& 0.3275 & 0.3732 & 0.0558 & 65
& 0.1473 & 0.2732 & 0.1140 & 35
& -- & -- & -- & 0 \\

\midrule
\multicolumn{13}{@{}l}{\textbf{Baseline}} \\
RNN-Adapt
& -0.1200 & 0.1415 & 0.0118 & 100
& -0.0744 & 0.1614 & 0.0256 & 100
& -0.0933 & 0.2082 & 0.0356 & 100 \\

Res-Adapt
& -0.1069 & 0.1421 & 0.0107 & 100
& -0.1291 & 0.1833 & 0.0156 & 100
& -0.1250 & 0.2062 & 0.0246 & 100 \\

\midrule
\multicolumn{13}{@{}l}{\textbf{Ours}} \\
\textbf{OSRAM}
& \textbf{-0.0059} & \textbf{0.0733} & 0.0078 & 100
& \textbf{0.0153} & \textbf{0.0929} & 0.0109 & 100
& \textbf{0.0279} & \textbf{0.1560} & 0.0155 & 100 \\

\bottomrule
\end{tabular}%
}
}

\vspace{3pt}
{\footnotesize
Comparison of OSRAM with ablation and baseline methods at different commanded velocities. Here, DA denotes dynamics adaptation, FD denotes forward dynamics, OL-Full denotes open-loop full-state dynamics, and OL-Out denotes open-loop output-state dynamics.}
\vspace{-15pt}
\end{table*}

\subsection{Why Model the Closed-Loop System?}

To address \textbf{Q2}, we examined whether modeling the closed-loop command-response behavior is more effective than relying on low-level action-conditioned dynamics models for adaptation. We compared the prediction accuracy and task level performance of OSRAM with two open-loop dynamics baselines under the same velocity-tracking policy.

The first baseline was an open-loop full-state dynamics model (\textbf{OL-Full}), which predicted the next state from full robot observations (e.g., linear and angular velocities, projected gravity, previous actions, and velocity commands in the velocity-tracking setting), together with the executed joint actions as control inputs. This setting corresponded to the standard partially observable open-loop dynamics formulation commonly used in model-based control.

The second baseline was an open-loop output-state dynamics model (\textbf{OL-Out}), which used the same reference-command-related observations as OSRAM's closed-loop dynamics model (e.g., base-frame $x$-$y$ linear velocity and $z$-axis angular velocity), along with joint actions as control inputs. This baseline removed information unrelated to the command while retaining an open-loop prediction structure.

We first compared the three models under identical training conditions: each model is pre-trained on the same amount of data and then finetuned on the same target dataset using the same batch size and number of gradient steps. As shown in Fig.~\ref{fig:method_comparison}(a), the closed-loop dynamics model achieved the highest prediction accuracy, while OL-Full and OL-Out yielded MAEs of $0.0789$ and $0.0338$, respectively.

More importantly, this difference persisted when the models are used for reference adaptation. We evaluated task-level performance on base velocity tracking tasks with different target velocities. Representative tracking results are shown in Fig.~\ref{fig:method_comparison}(b) and (c), and the corresponding tracking errors and success rates are summarized in Tab.~\ref{tab:velocity_tracking_comparison}. OSRAM achieved substantially better tracking performance across all tested velocities. At $0.4\,\mathrm{m/s}$ and $0.7\,\mathrm{m/s}$, OSRAM obtained much lower $E_{\mathrm{RMSE}}$ than OL-Full and OL-Out while maintaining a $100\%$ success rate. At the more challenging $0.9\,\mathrm{m/s}$ setting, both open-loop baselines failed with $0\%$ success rate, whereas OSRAM remained stable and achieved an $E_{\mathrm{RMSE}}$ of $0.1560$ with $100\%$ success rate. These results suggest that, under the same limited adaptation budget, directly modeling the task-level command-response behavior provides a more effective predictive representation than learning low-level action-conditioned dynamics.






\subsection{Comparison with Online Adaptation Strategies}

To address \textbf{Q3}, we compared the proposed method with representative online adaptation-based sim-to-real transfer baselines in simulation using velocity-tracking tasks. The first baseline, denoted as RNN Adaptation (\textbf{RNN-Adapt}), followed an RMA-style~\cite{kumar2021rma} adaptation procedure, in which a recurrent neural network (RNN) encoder took a history of observations as input and encoded it into a latent representation provided to the velocity-tracking policy. The second baseline, denoted as Residual Policy Adaptation (\textbf{Res-Adapt}), followed a residual learning-based adaptation setup~\cite{loco_residual}. For this baseline, the base policy was kept frozen while a residual policy was learned on top of its actions. For a fair comparison, the residual policy was finetuned using the same amount of target-domain data used to finetune the closed-loop dynamics model in OSRAM. The data was collected from the same distribution as the testing environment.

The Baseline section of Tab.~\ref{tab:velocity_tracking_comparison} summarizes the tracking performance of the two online adaptation baselines. Both RNN-Adapt and Res-Adapt maintained a $100\%$ success rate across all tested velocities, indicating stable execution under the evaluated conditions. However, their tracking accuracy was consistently lower than that of OSRAM. In particular, RNN-Adapt obtained $E_{\mathrm{RMSE}}$ values of $0.1415$, $0.1614$, and $0.2082$ at $0.4$, $0.7$, and $0.9\,\mathrm{m/s}$, respectively, while Res-Adapt obtained $0.1421$, $0.1833$, and $0.2062$. In comparison, OSRAM achieved lower $E_{\mathrm{RMSE}}$ values of $0.0733$, $0.0929$, and $0.1560$ under the same velocity settings. OSRAM also achieved the best $E_{\mathrm{mean}}$ across all velocities, indicating more accurate average velocity tracking error mitigation. Thus, under our evaluation setting, adapting through the reference interface provides more accurate tracking than the representative latent and residual policy-adaptation strategies.

\begin{table}[t]
\centering
\caption{Hardware Velocity Tracking at $v_x = 0.7\,\mathrm{m/s}$}
\label{tab:hardware_velocity_tracking_comparison}
{
\scriptsize
\setlength{\tabcolsep}{3.2pt} 
\renewcommand{\arraystretch}{0.96}
\resizebox{\columnwidth}{!}{%
\begin{tabular}{@{}lcccc@{}}
\toprule
\textbf{Method}
& $E_{\mathrm{mean}}$
& $E_{\mathrm{RMSE}}$
& $E_{\mathrm{std}}$
& $SR(\%)$ \\
\midrule

OSRAM w/o DA
&-0.2298  &0.2847  &0.0318 &100 \\

RNN-Adapt
&-0.1354  &0.2109  &0.0064  &100  \\

OSRAM
&-0.0827  &0.2014  &0.0169  &100  \\

OSRAM w/ RNN-Adapt
&-0.1023  &0.1962  &0.0066  &100 \\

\bottomrule
\end{tabular}%
}
}

\vspace{4pt}
{\footnotesize
Comparison of OSRAM with baseline methods for hardware velocity tracking at $v_x = 0.7\,\mathrm{m/s}$. DA denotes dynamics adaptation, and OSRAM w/ RNN-Adapt denotes replacing the baseline policy with an RNN-based adaptation policy.}
\vspace{-20pt}
\end{table}

\subsection{Sim-to-Real Hardware Evaluation}

To address \textbf{Q4}, we further evaluated OSRAM in two real-world sim-to-real transfer scenarios covering different task objectives, policy configurations, command-response spaces, and hardware configurations. The experiments assessed three aspects of OSRAM: 1)\ whether it enables real-world sim-to-real transfer, 2)\ whether it applies to different task objectives, policy configurations, and command-response spaces, and 3)\ whether it can complement existing online adaptation mechanisms.

\begin{figure}[t]
    \centering
    \includegraphics[width=\columnwidth]{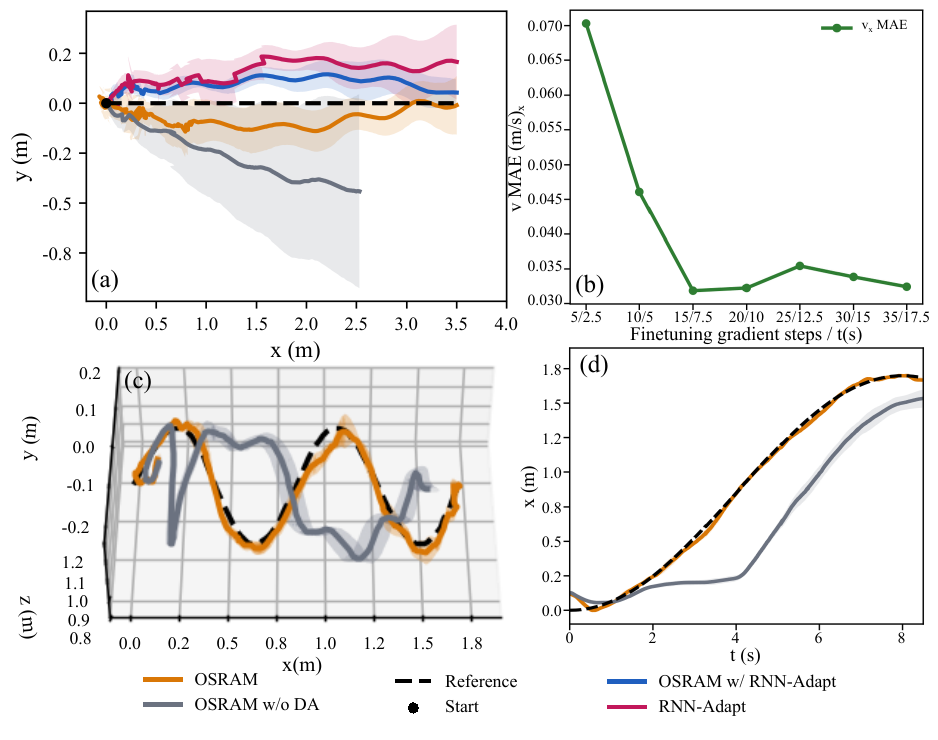}
    \vspace{-20pt}
    \caption{Hardware results of velocity-tracking and loco-manipulation. (a) Global position trajectories for velocity tracking, truncated at $x=3.5\,\mathrm{m}$ to highlight lateral tracking error. (b) Forward-velocity prediction error, measured by $v_x$ MAE, versus the number of finetuning gradient steps. (c) End-effector global position trajectories for loco-manipulation. (d) Global $x$-position trajectories for loco-manipulation.}
    \label{fig:hardware_results}
    \vspace{-10pt}
\end{figure}

\noindent\textbf{Velocity tracking.} We first evaluated OSRAM on hardware using velocity-tracking tasks. We compared four configurations: a baseline policy, the baseline policy augmented with OSRAM, an RNN-Adapt policy, and the RNN-Adapt policy augmented with OSRAM. Improvements over the baseline policy showed that the proposed method can be applied to a standard control policy, while additional gains over the RNN-based policy indicate that OSRAM could further improve policies that already included online adaptation. The resulting trajectory tracking results are reported in Tab.~\ref{tab:hardware_velocity_tracking_comparison}. Position tracking in the sagittal and coronal planes is shown in Fig.~\ref{fig:hardware_results}(a), where methods augmented with OSRAM improved coronal tracking performance for both baseline policy and RNN-Adapt policy. To assess how the number of fine-tuning gradient steps affects the closed-loop dynamics model, we plotted the forward-velocity prediction MAE as a function of gradient steps. The MAE decreased rapidly during the initial fine-tuning stages and then plateaued after approximately $30$ steps.

\noindent\textbf{Loco-manipulation.} We next evaluated OSRAM on a loco-manipulation end-effector position-tracking task, in which the robot was commanded to track a global sine-wave end-effector trajectory. Because global position command-response signals lie in an infinite Euclidean space, the meta-dynamics inputs and outputs are represented in the robot’s local base frame. The policy design and training recipe followed prior work~\cite{hiwet}. The resulting trajectory tracking performance and cumulative tracking errors over the trajectory evolution are shown in Fig.~\ref{fig:hardware_results}(c) and 4(d) where OSRAM reduced the \(x\)-direction RMSE from \(36.94 \pm 3.02\) cm to \(3.11 \pm 0.31\) cm and the \(y\)-direction RMSE from \(3.48 \pm 0.45\) cm to \(2.72 \pm 0.23\) cm. The trajectories in Fig.~\ref{fig:hardware_results}(c)--(d) show that OSRAM substantially compensates for the large systematic tracking error present without adaptation.


\subsection{How Does Reference Adaptation Correct Tracking Error?}

\begin{figure}[t]
    \centering
    \includegraphics[width=\columnwidth]{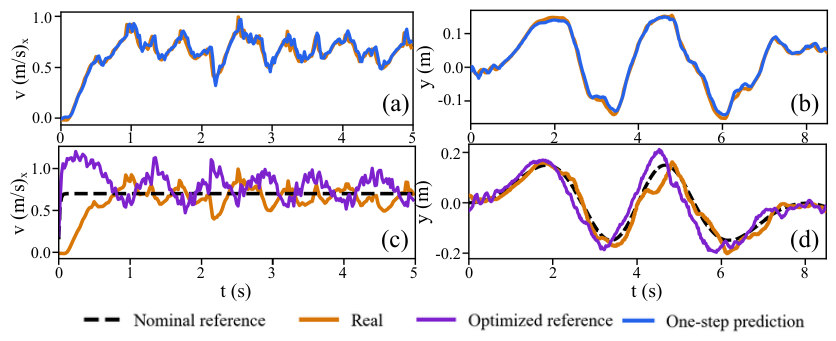}
    \vspace{-20pt}
    \caption{(a)--(b) One-step predictions of the finetuned closed-loop dynamics model for velocity tracking and loco-manipulation, respectively. (c)--(d) Nominal references, OSRAM-adapted references, and realized responses for forward-velocity tracking and end-effector $y$-position tracking. The adapted references compensate for systematic tracking errors so that the realized responses more closely follow the nominal targets.}
    \label{fig:analysis}
    \vspace{-20pt}
\end{figure}

Finally, we examine the mechanism how OSRAM improves tracking through the interaction between closed-loop prediction and reference adaptation. Fig.~\ref{fig:analysis}(a)--(b) compares one-step predictions from the finetuned closed-loop model with the realized responses for velocity tracking and loco-manipulation. The model achieves prediction RMSEs of $0.0557\,\mathrm{m/s}$ and $0.0074\,\mathrm{m}$, respectively, showing that it captures the deployed system's command-response behavior with sufficient accuracy for reference optimization.

Fig.~\ref{fig:analysis}(c)--(d) illustrates how these predictions translate into corrective reference commands. Instead of attempting to make the robot directly follow the nominal reference, OSRAM selected a modified reference whose predicted response more closely matched the nominal target. For example, when the robot systematically under-tracked a command, OSRAM increased the reference to compensate. When it overshot, OSRAM reduced the reference. Consequently, the realized trajectories moved closer to the original desired trajectories. These results illustrate the central mechanism of OSRAM: learning the residual command-response behavior of the deployed closed-loop system enables compensation of tracking errors through the reference interface while leaving the underlying policy unchanged.

\section{Conclusion, Limitation And Future Work}
\label{sec:conclusion}
We presented OSRAM, an online sim-to-real adaptation framework that models the deployed robot and its controller as a closed-loop command-response system and corrects residual tracking errors by adapting reference commands without modifying the policy. Across simulated and physical bipedal velocity tracking and loco-manipulation, OSRAM improves prediction and tracking in unseen dynamics.

While we focus on relatively low-dimensional command spaces, an important future direction is to extend OSRAM to high-dimensional references, such as whole-body motion trajectories. One possible approach is to compress reference motions into a low-dimensional latent representation and perform adaptation in that space. In addition, the current sampling-based optimization backend can be sensitive to hyperparameter tuning. Since the learned closed-loop dynamics model is differentiable, future work could explore gradient-based optimization as a more efficient alternative. Finally, because OSRAM does not assume a specific form of the inner-loop control policy, extending the proposed framework to model-based controllers can further broaden its applicability.

\bibliographystyle{IEEEtran}
\bibliography{example}

\vfill

\end{document}